\documentclass{llncs}
\usepackage{amssymb}
\usepackage{amsmath}

\newtheorem{assumption}{Assumption}
\newtheorem{myremark}{Remark}

\newcommand{\nR}{\mathbb{R}}

\newcommand{\nL}{\mathcal{L}}
\newcommand{\regret}{\mathcal{R}}
\newcommand{\ra}{\rightarrow}

\newcommand{\PP}[1]{\mathbb{P}\left[ #1 \right] }

\newcommand{\EE}[1]{\mathbb{E}\left[ #1 \right] }

\newcommand{\Ind}[1]{\mathbb{I}\left[ #1 \right] }

\newcommand{\bx}{\mathbf{x}}
\newcommand{\bz}{\mathbf{z}}

\newcommand{\hmu}{\widehat{\mu}}
\newcommand{\hsigma}{\widehat{\sigma}}
\newcommand{\invphi}{\varphi^{-1}}
 
\newcommand{\argmin}{\operatornamewithlimits{argmin}}

\newcommand\litem[1]{\item{\bfseries #1:}}

\begin{document}

\title{Generalized Risk-Aversion \\ in Stochastic Multi-Armed Bandits}
\titlerunning{Risk-Aversion in Multi-Armed Bandits}

\author{Alexander Zimin \and Rasmus Ibsen-Jensen \and Krishnendu Chatterjee}
\institute{Institute of Science and Technology Austria\\\email{\{azimin,ribsen,krishnendu.chatterjee\}@ist.ac.at}}

\maketitle

\begin{abstract}
We consider the problem of minimizing the regret in stochastic multi-armed bandit, when the measure of goodness of an arm is not the mean return, but some general function of the mean and the variance. We characterize the conditions under which learning is possible and present examples for which no natural algorithm can achieve sublinear regret.
\end{abstract}

\section{Introduction}\label{section:intro}

The stochastic multi-armed bandit problem is a well-studied framework to model sequential decision-making problems. It has a wide range of theoretical as well as practical applications such as clinical trials, web advertisement placement, packet routing, to name a few. In the usual formulation, an agent (a learner, or an algorithm) has to choose from one of several unknown distributions (which are called arms), receive a sample (a loss) from the arm chosen, and repeat this process for some prescribed amount of time. The goal of the learner is expected regret minimization, i.e., minimization of the expectation of the difference between its own cumulative loss and the cumulative loss of the best arm, where the best arm is the one with the smallest mean. However, for some applications the expected criterion might not be the most desirable. For example, in clinical trials one might not be interested in the most effective treatment on average, but in the one that is more robust and still has a good effect on average. In terms of multi-armed bandits, in this case the best arm is defined not by the mean, but by some risk measure, which is a function of the distribution itself. This leads to the idea of the risk-averse bandit problem.

Risk-aversion has been extensively studied in other fields. Starting from the economic theory (\cite{Markowitz01}, \cite{Neumann01}) and ending up with the neighbouring field of reinforcement learning (\cite{Defourny01}, \cite{Shen01}, \cite{Shen02}, \cite{Patek01}). In the field of online learning, risk-aversion was studied in the experts setting by \cite{Even-dar03}. They obtained several negative and positive results for when Sharpe-ratio (\cite{Sharpe01}) and  mean-variance (\cite{Markowitz01}) was used as risk measures. \cite{Warmuth01} studied the problem of pure variance minimization.
Other risk measures were studied in \cite{Sani01} and \cite{Maillard01}. The former proposes to use the mean-variance criterion as a measure of risk and aims at minimizing the notion of the regret that takes into account the variability of the algorithm. The latter considers log-exponential risk measure, which belongs to the class of so-called coherent risk measures (\cite{Rockafellar03}) and minimizes the regret defined using this measure.

There is no universally agreed notion of what a good measure of risk is, and the appropriate notion can vary from one problem to another. All previous works focused on some particular risk measures, which has immediately limited the applicability of the results and raised a lot of questions on the quality of the particular risk measure. In this work, we consider a different approach: instead of a specific risk measure, we define the risk-averse bandit problem with arbitrary (but fixed) risk measure and the corresponding regret. We focus on risk measures defined as a function of the first two moments (the mean and the variance). 
This generalizes the setting of \cite{Sani01} from linear to arbitrary functions, while considering notion of regret similar to \cite{Maillard01}.

We present two motivating examples of our framework: (1)~We consider the \emph{threshold variance} problem, where we have the usual bandit setting and interested in the means of the distributions (of the arms), but would like to chose only from those arms that has the variance smaller than a specified threshold. One possible formalization of this problem leads us to the risk-averse regret minimization with discontinuous function of the mean and the variance used as a risk measure. (2)~Consider a risk measure that is a linear combination of the mean and the square root of the variance, where both the summands are of the same order. This is a natural variant of the mean-variance optimization and is a continuous function of the mean and the variance.

Our main results are as follows: (1)~First we present an algorithm, namely, $\varphi$-LCB, which belongs to the wide family of Lower (Upper) Confidence Bound algorithm (the descendants of UCB algorithm of \cite{Auer02}, see also, e.g. \cite{Audibert01}, \cite{Garivier01}), 
and prove logarithmic risk-averse regret bounds for all continuous functions. (2)~Second, we present an example of a discontinuous function where no natural algorithm (based on the optimism in face of uncertainty principle) can achieve 
sublinear regret. (3)~Finally, we present another algorithm, namely, $\varphi$-LCB2, that makes learning feasible with the mild assumption that no arm hits the discontinuity points.
Our proof approach is similar to \cite{Sani01} and \cite{Maillard01}, while the latter used slightly different KL-divergence based version of the algorithm (\cite{Maillard02}).

\smallskip\noindent{\em Other related works.}
In the bandit setting risk-aversion has been approached from different perspectives. \cite{Galichet01} designs an algorithm that uses conditional value at risk (CVaR) as a risk measure. However, they aim at minimizing the usual expected regret under the assumption that the best mean arm is also the best risk-aversion arm, which is completely different from our goal. \cite{Yu04} derive PAC-bounds on the single- and multi-period risk for several different risk measures, nevertheless, the PAC-style of their results makes it inapplicable to our problem. \cite{Salomon01} considers the deviations of the regret in the standard setting, which seem to address the same issues, but it remains unclear if their results can be connected to risk-averse regret minimization.

\smallskip\noindent{\em Organization.} In Section~\ref{section:problem} we introduce the notations to be used, formally state the problem, and present some examples which can be modeled in our framework. In Sections~\ref{section:continuous} and \ref{section:discontinuous} we discuss two cases of the main problem and present the corresponding algorithms together with the risk-averse regret bounds. Section~\ref{section:discussion} discusses open problems and the possible extensions of the setting. The paper concludes with the proofs of the main theorems in Section~\ref{section:proofs}.

\section{The problem}\label{section:problem}
Let $\nL_2$ denote the set of distributions supported on $\left[ 0,1\right]$. We consider the stochastic multi-armed bandit setting with $K$ arms and $\nu_1, .., \nu_K \in \nL_2$ being the distributions of arms. At time step $t$ the learner chooses arm $a_t$ to pull and receives a sample $X_{a_t,T_{a_t}(t)}$ drawn from $\nu_{a_t}$, where $T_i(t)$ is the number of times that arm $i$ is pulled by the $t$-th time step, that is,
\[
T_i(t) = \sum_{s=1}^{t}\Ind{a_s = i}\enspace .
\]
We consider the case where the learner is given a risk measure $R:\nL_2\ra\nR$. The risk measure of arm $i$ is $R_i = R(\nu_i)$. This measure defines the best arm $i^\star$ by
\[
i^\star = \argmin_{i=1..K}R_i
\]
and the goal of the algorithm is to identify that arm.
The performance of the algorithm is measured by means of risk-averse regret:
\[
\regret_n = \sum_{t=1}^{n}R_{a_t} - \sum_{t=1}^{n}R_{i^\star} = \sum_{t=1}^{n}R_{a_t} - n \cdot R_{i^\star}\enspace .
\]
Note that this corresponds to the notion of pseudo-regret for stochastic bandits, but there is no regret notion in our setting that directly corresponds to true regret in stochastic bandits. 
One could try to define true regret as the difference of risk measures applied to the empirical distributions of the algorithm and the best arm (similar to~\cite{Sani01}). 
However, then the algorithm could be punished even for switching between the best arms, 
which can be an undesirable feature. 

Some examples of such risk measures are $R(X) = \EE{X}$ with $X$ being a random variable (usual stochastic bandit) and $R(X) = \frac{1}{\lambda}\log{\EE{\exp{\lambda X}}}$, considered in \cite{Maillard01}.

In this paper we focus on the risk measures of the following form:
\[
R(X) = f(\EE{X}, \textrm{Var}(X))\enspace.
\]
In other words, the learner is supplied by a function $f: D \ra \nR$, where $D = \left[0,1\right]\times \left[0, 1 \right]$\footnote{The domain of the second argument can be restricted to $\left[0, \frac{1}{4} \right]$, since for a random variable which takes values in $\left[0,1\right]$, the variance is upper bounded by $\frac{1}{4}$.}. If we denote the risk measure of arm $i$ by $f_i$, i.e. $f_i = f(\mu_i,\sigma^2_i)$, where $\mu_i$ and $\sigma^2_i$ are the mean and the variance of the $i$-th arm respectively, then $i^\star = \argmin_{i=1..K}{f_i}$ and the regret is
\[
\regret_n = \sum_{t=1}^{n}f_{a_t} - \sum_{t=1}^{n}f_{i^\star} = \sum_{t=1}^{n}f_{a_t} - n \cdot f_{i^\star}\enspace .
\]
Our class of risk measures is rich enough to model a lot of interesting problems:
\begin{enumerate}
\litem{Standard Bandit} $f(x,y) = x$. This is the standard stochastic multi-armed bandit setting.
\litem{Variance Minimization} $f(x,y) = y$. This is the variance minimization problem, considered in \cite{Warmuth01}.
\litem{Mean-variance Bandit} $f(x,y) = x + \lambda \cdot  y$. This is a version of the problem considered in \cite{Sani01}. 
A related and natural variant is $f(x,y) = x + \lambda \sqrt{y}$, where both summands are of the same order. 
\litem{Threshold Variance} $f(x,y) = x\Ind{y < v} + \Ind{y \geq v}$. This risk measure can be used to model threshold variance problem described in Section~\ref{section:intro}.
\litem{Log-Exponential Risk} $f(x,y) = x + \frac{\lambda}{2}x^2 + \frac{\lambda}{2}y$. This measure can be seen as an approximation to the coherent risk measure, considered in \cite{Maillard01}: $\frac{1}{\lambda}\log{\EE{\exp{\lambda X}}}$, when it is restricted to the first two moments.
\end{enumerate}
Our goal is to study conditions on the function $f$ under which learning is possible.

\section{Our Results}\label{section:results}
We distinguish between two cases of the problem: continuous and discontinuous functions $f$. In the continuous case we prove that learning is possible for every function. In the discontinuous case we present an example where learning is not possible.
The negative example motivates a restriction, and we show that under the restriction 
learning is feasible.

\subsection{Continuous functions}\label{section:continuous}
In this section we will show that learning is possible for any continuous function $f$. 
We start with a characterization of continuous functions that will be used to present the algorithm.

\begin{lemma}\label{lemma:continuous}
For every continuous function $f: D \ra \nR$, there exists a function $\varphi: \nR_+ \ra \nR_+$, such that
\begin{enumerate}
\item $ \varphi(0) = 0$;
\item $\varphi$ is a strictly increasing function;
\item $|f(\bx_2)-f(\bx_1)| \leq \varphi(||\bx_2-\bx_1||_1) \textrm{ for all } \bx_1, \bx_2 \in D$.
\end{enumerate}
\end{lemma}
As an example, consider an $\alpha$-H\"{o}lder continuous function $f$: in this case $\varphi(z) = cz^\alpha$ would satisfy the conditions of Lemma~\ref{lemma:continuous} by the definition of  $\alpha$-H\"{o}lder continuity. But Lemma~\ref{lemma:continuous} is stated for every continuous function: as another example, consider the continuous function
\begin{equation}\label{not_holder_func}
h(x) = \begin{cases}
 \frac{-1}{\ln(x/2)} &\mbox{ if $x\in D$ and $x > 0$} \\
  0 &\mbox{ if $x=0$ \enspace .}
       \end{cases}
\end{equation}
It is not $\alpha$-H\"{o}lder continuous for any $\alpha$, but $\varphi(z) = h(z)$ satisfies the conditions of Lemma~\ref{lemma:continuous} for $f(x,y) = h(x)$.

We will use Lemma \ref{lemma:continuous} to construct a high-confidence interval for $f$ from the confidence intervals for its arguments. We start by defining the empirical mean and the empirical variance of arm $i$:
\[
\hmu_{i,t} = \frac{1}{t}\sum_{s=1}^{t}X_{i,s} \textrm{\,\, and \,\,} \hsigma^2_{i,t} = \frac{1}{t}\sum_{s=1}^{t}(X_{i,s} - \hmu_{i,t})^2\enspace .
\]
The following concentration results are the basis for our argument.
\begin{lemma}[Chernoff-Hoeffding bound]\label{lemma:chernoff}
For every $i=1,\dots,K$, $t=1,\dots,n$, and $\delta \in (0,\frac{1}{2})$, with probability at least $1-2\delta$
\[
|\hmu_{i,t}-\mu_i| \leq \sqrt{\frac{\ln\frac{1}{\delta}}{2t}}\enspace .
\]
\end{lemma}
\begin{lemma}[Lemma 2 from \cite{Antos01}]\label{lemma:antos}
For all $i=1,\dots,K$, $t=1,\dots,n$, and $\delta \in (0,\frac{1}{4Kn})$, with probability at least $1-4Kn\delta$
\begin{equation}\label{bound:deviation}
|\hsigma^2_{i,t} - \sigma^2_i| \leq 5\sqrt{\frac{\ln\frac{1}{\delta}}{2t}} \enspace .
\end{equation}
\end{lemma}
From Lemma~\ref{lemma:continuous}, Lemma~\ref{lemma:chernoff}, and Lemma~\ref{lemma:antos} we can construct the following high-confidence bound for $f$:
\begin{equation}\label{bound:risk}
|f(\hmu_{i,t}, \hsigma^2_{i,t}) - f_i| \leq \varphi \left( 6\sqrt{\frac{\ln\frac{1}{\delta}}{2t}}\right)\enspace .
\end{equation}
The algorithm $\varphi$-LCB will at time step $t$ choose an arm that minimizes the corresponding lower confidence bound:
\begin{equation}\label{decision_rule}
a_t = \argmin_{i=1..K}\left[ f(\hmu_{i,T_i(t-1)}, \hsigma^2_{i,T_i(t-1)}) - \varphi\left(6\sqrt{\frac{\ln\frac{1}{\delta}}{2\cdot T_i(t-1)}}\right) \right]\enspace .
\end{equation}
The algorithm chooses arm $i$ if $f(\hmu_{i,T_i(t-1)}, \hsigma^2_{i,T_i(t-1)})$ is really small or if $\varphi\left(6\sqrt{\frac{\ln\frac{1}{\delta}}{2\cdot T_i(t-1)}}\right)$ is big. The former means that the algorithm tries to exploit the arm that has small estimated risk measures, while the latter means that the estimate for the arm $i$ is rough and the algorithm tries to improve it by exploring this arm further. In other words, the $\varphi$-LCB algorithm tries to deal with exploration-exploitation trade-off using the so-called optimism in face of uncertainty principle.
\begin{figure}[h]
\centering
\fbox{
\begin{minipage}{.95\textwidth}
{\bfseries Parameters}: Confidence level $\delta$;\\
{\bfseries For all time steps $t=1,2,\dots,n$, repeat}
\begin{enumerate}
\item Compute $a_t = \argmin_{i=1..K}\left[ f(\hmu_{i,T_i(t-1)}, \hsigma^2_{i,T_i(t-1)}) - \varphi\left(6\sqrt{\frac{\ln\frac{1}{\delta}}{2\cdot T_i(t-1)}}\right) \right]$.
\item Output $a_t$ as a decision.
\item Receive $X_{a_t,T_{a_t}(t)} \sim \nu_{a_t}$.
\end{enumerate}
\end{minipage}
}
\caption{The $\varphi$-LCB algorithm}
\label{fig:phi-ucb}
\end{figure}

Theorem~\ref{theorem:continuous} states the regret bound of the $\varphi$-LCB algorithm.

\begin{theorem}[Feasibility of learning]\label{theorem:continuous}
Consider a continuous function $f$, then for $\delta\in(0,\frac{1}{4Kn})$ with probability at least $1-4Kn\delta$ the regret of the $\varphi$-LCB algorithm at time $n$ is upper bounded by:
\[
\regret_n \leq \sum_{i: \Delta_i > 0} \frac{18\cdot \Delta_i \cdot \ln\frac{1}{\delta}}{(\invphi(\Delta_i/2))^2} + \sum_{i: \Delta_i > 0} \Delta_i\enspace ,
\]
where $\Delta_i = f_i - f_{i^\star}$. Moreover, for $n > 4K$, if the algorithm is run with $\delta = \frac{1}{n^2}$, then with probability at least $1-\frac{4K}{n}$ the regret is 
upper bounded by:
\[
\regret_n \leq \sum_{i: \Delta_i > 0} \frac{36\cdot \Delta_i}{(\invphi(\Delta_i/2))^2}\ln{n} + \sum_{i: \Delta_i > 0} \Delta_i\enspace .
\]
\end{theorem}

\noindent{\bf Efficiency.} Theorem~\ref{theorem:continuous} shows that learning is feasible for every continuous function. 
We now discuss the efficiency of the algorithm with respect to different classes of continuous functions.
\begin{enumerate}

\litem{Lipschitz functions} when $f$ is $L$-Lipschitz, i.e. $\varphi(z) = Lz$, the regret bound is
\[
\regret_n \leq \sum_{i: \Delta_i > 0} \frac{144\cdot L^2}{\Delta_i}\ln{n} + \sum_{i: \Delta_i > 0} \Delta_i
\]
and the dependence on $\Delta_i$ in front of $\ln{n}$ matches the dependence in the regret of the $\varphi$-LCB algorithm in the standard stochastic bandit problem. 
The worse constant ($144 L^2$) term is an artifact of doing such general analysis. This case covers the standard bandit and the variance minimization problems with $L=1$, the log-exponential risk problem with $L = 1 + \lambda$, and the mean-variance bandit problem with $f(x,y) = x + \lambda y$ in which  $L=\max\left\lbrace 1,\lambda\right\rbrace $.

\litem{H\"{o}lder functions} when $f$ is $\alpha$-H\"{o}lder continuous, i.e. $\varphi(z) = Lz^\alpha$, the regret bound is
\[
\regret_n \leq \sum_{i: \Delta_i > 0} \frac{36\cdot (2\cdot L)^{\frac{2}{\alpha}}}{(\Delta_i)^{\frac{2-\alpha}{\alpha}}}\ln{n} + \sum_{i: \Delta_i > 0} \Delta_i\enspace .
\]
This case covers the mean-variance problem with $f(x,y) = x + \lambda \sqrt{y}$ which is $\frac{1}{2}$-H\"{o}lder continuous with $L=\max\left\lbrace1,\lambda\right\rbrace $. Note that the dependence on $\Delta_i$ in this case is worse than for Lipschitz functions, but it is still polynomial.

\litem{Non-H\"{o}lder functions} to demonstrate how efficiency can decrease for the general class of continuous functions, consider $f(x,y) = h(x)$ from (\ref{not_holder_func}), then $\varphi(z) = h(z)$ and the regret bound becomes
\[
\regret_n \leq \sum_{i: \Delta_i > 0} 9\cdot \Delta_i \cdot e^{4/\Delta_i}\ln{n} + \sum_{i: \Delta_i > 0} \Delta_i\enspace .
\]
We can see that the term in front of $\ln{n}$ grows exponentially as $\Delta_i$ goes to $0$ in comparison to the polynomial growth for Lipschitz and H\"{o}lder functions.
\end{enumerate}
\begin{myremark}
Note that it is possible to design an anytime version of $\varphi$-LCB for the case when $n$ is not known in advance. To do so, at each time step we take $\delta = \varepsilon_t$, where $\varepsilon_t$ is a sequence decreasing at an appropriate rate. However, we do not pursue this direction further.
\end{myremark}

\subsection{Discontinuous functions}\label{section:discontinuous}
The case of discontinuous functions is more tricky. We present a negative example and a partially positive result.
We start with an example of a discontinuous function $f$ where no algorithm following the 
optimism in face of uncertainty principle can achieve sublinear regret.

\begin{example}
Consider the following discontinuous function:
Let 
\[
f(x,y) = \begin{cases}
 1 &\mbox{ if $x = 0.5$ and $y = 0.1$;} \\
 \frac{1}{2} &\mbox{ if $y\geq 0.5$;} \\  
 0 &\mbox{ otherwise}\enspace .
       \end{cases}
\]
Consider two arms $1$ and $2$ such that 
$\mu_1=0.5$ and $\sigma^2_1=0.1$ and 
$\mu_2=1$ and $\sigma^2_1=0.75$.
Then any algorithm based on the optimism in face of uncertainty principle will keep on choosing arm~1
with non-negligible probability.
This is because if the estimate of the algorithm is not precisely the 
discontinuity point, then arm~1 will be chosen due to optimism.  
\end{example}

However, in the case when no arm hits the discontinuity point, learning is possible as we will show. Let $d_i(x,y) = |x-\mu_i|+|y-\sigma_i^2|$ be the distance to the point representing $i$-th arm. Define $\Omega_f$ to be the set of discontinuities of $f$ and $d_\Omega(x,y) = \inf_{(z_1,z_2)\in\Omega_f}\left\lbrace |z_1-x| + |z_2-y| \right\rbrace$ to be the distance to the closest discontinuity point. We will show that learning is possible under the following assumption.

\begin{assumption}\label{assumption1}
For each arm $i$ there exists $\varepsilon > 0$ such that $f$ is continuous in $B_i(\varepsilon) = \left\lbrace (x,y) \in D : d_i(x,y) \leq \varepsilon \right\rbrace $.
\end{assumption}
Let us introduce $e_i = \sup\left\lbrace \varepsilon>0 : f \textrm{ is continuous in } B_i(\varepsilon) \right\rbrace = d_\Omega(\mu_i,\sigma_i^2)$, then by Lemma~\ref{lemma:continuous} there exists a function $\varphi_i$ that satisfies the required condition, but only in $B_i(e_i)$ instead of $D$. So when our estimated values are in $B_i(e_i)$ we can use the same algorithm as before. We present a new algorithm $\varphi$-LCB2 that first pulls each arm some amount of times, such that with high probability $(\hmu_{i,t},\hsigma^2_{i,t})$ is in $B_i(e_i)$ for each arm, in other words, that $d_i(\hmu_{i,t},\hsigma^2_{i,t}) \leq e_i$. If we would know $e_i$ in advance, then to ensure this condition with high probability it is enough (from Lemma \ref{lemma:chernoff} and Lemma \ref{lemma:antos}) that
\[
6\sqrt{\frac{\ln\frac{1}{\delta}}{2t}} \leq e_i\enspace .
\]
Hence, we would need to pull each arm $18e_i^{-2}\ln\frac{1}{\delta}$ times. But since $e_i$ is not known in advance, we would pull each arm until its distance to $(\mu_i,\sigma_i)$ is twice less than distance to the closest discontinuity point. Formally, the algorithm chooses each arm until
\begin{equation}\label{stopping_theory}
d_i(\hmu_{i,t},\hsigma^2_{i,t}) \leq \frac{1}{2}d_\Omega(\hmu_{i,t},\hsigma^2_{i,t})\enspace .
\end{equation}

 At the time when this happens, we can be sure that  $(\hmu_{i,t},\hsigma^2_{i,t}) \in B_i(e_i)$ and this procedure does not increase the number of pulls too much. To ensure (\ref{stopping_theory}) with high probability it is enough that
\begin{equation}\label{stopping}
6\sqrt{\frac{\ln\frac{1}{\delta}}{2t}} \leq \frac{1}{2}d_\Omega(\hmu_{i,t},\hsigma^2_{i,t})\enspace .
\end{equation}
After ensuring this for each arm, the algorithm proceeds as the $\varphi$-LCB algorithm, but uses $\varphi_i$ for each arm instead of a common function $\varphi$:
\begin{equation}\label{decision_rule2}
a_t = \argmin_{i=1..K}\left[ f(\hmu_{i,T_i(t-1)}, \hsigma^2_{i,T_i(t-1)}) - \varphi_i\left(6\sqrt{\frac{\ln\frac{1}{\delta}}{2\cdot T_i(t-1)}}\right) \right]\enspace .
\end{equation}
Note that constructing $\varphi_i$ requires knowledge of $e_i$, but this can also be avoided if we construct it in the estimated (and smaller) region, defined at the time, when (\ref{stopping}) occurs.
The following theorem states the regret bound of the resulting algorithm.
\begin{figure}[h]
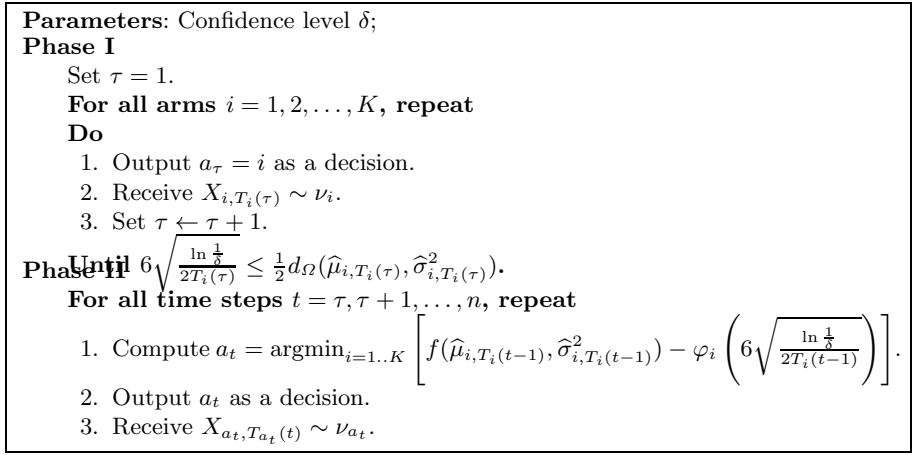

\centering
\fbox{
\begin{minipage}{.95\textwidth}
{\bfseries Parameters}: Confidence level $\delta$;\\
\begin{description}

\vspace{-7mm}
\item[Phase I] \hfill \\
Set $\tau = 1$.\\
{\bfseries For all arms $i=1,2,\dots,K$, repeat}\\
{\bfseries Do}
\begin{enumerate}
\item Output $a_\tau = i$ as a decision.
\item Receive $X_{i,T_{i}(\tau)} \sim \nu_{i}$.
\item Set $\tau \leftarrow \tau + 1$.
\end{enumerate}
{\bfseries Until $6\sqrt{\frac{\ln\frac{1}{\delta}}{2T_i(\tau)}} \leq \frac{1}{2}d_\Omega(\hmu_{i,T_i(\tau)},\hsigma^2_{i,T_i(\tau)})$.}\\

\vspace{-7mm}
\item[Phase II] \hfill \\
{\bfseries For all time steps $t=\tau,\tau+1,\dots,n$, repeat}
\begin{enumerate}
\item Compute $a_t = \argmin_{i=1..K}\left[ f(\hmu_{i,T_i(t-1)}, \hsigma^2_{i,T_i(t-1)}) - \varphi_i\left(6\sqrt{\frac{\ln\frac{1}{\delta}}{2T_i(t-1)}}\right) \right]$.
\item Output $a_t$ as a decision.
\item Receive $X_{a_t,T_{a_t}(t)} \sim \nu_{a_t}$.
\end{enumerate}

\end{description}

\end{minipage}
}
\caption{$\varphi$-LCB2 algorithm}
\label{fig:phi-ucb2}
\end{figure}
\begin{theorem}\label{theorem:discontinuous}
Consider function $f$ that satisfies Assumption \ref{assumption1}.
Then for $\delta\in (0,\frac{1}{4Kn})$ with probability at least $1-4Kn\delta$ for all $n \geq \sum_{i=1..K} 162\cdot e_i^{-2}\ln\frac{1}{\delta}$ the regret of the $\varphi$-LCB2 algorithm at time $n$ is upper bounded by:
\[
\regret_n \leq \sum_{i: \Delta_i > 0}\Delta_i\left( 162e_i^{-2} + \frac{18}{(\invphi_i(\Delta_i/2))^2} \right) \ln\frac{1}{\delta} + \sum_{i: \Delta_i > 0}\Delta_i
\]
where $\Delta_i$ and $e_i$ as defined before. Moreover, if the algorithm is run with $\delta = \frac{1}{n^2}$, then with probability at least $1-\frac{4K}{n}$ for all $n \geq \sum_{i=1..K} 324\cdot e_i^{-2}\ln n$
the regret is upper bounded by: 
\[
\regret_n \leq \sum_{i: \Delta_i > 0}\Delta_i\left( 324e_i^{-2} + \frac{36}{(\invphi_i(\Delta_i/2))^2} \right) \ln{n} + \sum_{i: \Delta_i > 0}\Delta_i\enspace .
\]
\end{theorem}
The theorem can be applied to our motivating example: the threshold variance problem. There are two continuous regions, when $y < v$ and when $y \geq v$. In either case we can take $\varphi(z) = z$ (in fact, we can take any increasing function for the region $y \geq v$, since $f$ is just a constant there) and then the bound becomes
\[
\regret_n \leq \sum_{i: \Delta_i > 0}4\left( 81\cdot e_i^{-2}\Delta_i + \frac{36}{\Delta_i} \right) \ln{n} + \sum_{i: \Delta_i > 0}\Delta_i\enspace .
\]
Actually, in this case the bound can be improved, since after Phase I the algorithm would know which arms have variance greater than $v$ and it would not pull them at all. Hence, for such arms term $4\frac{36}{\Delta_i}\ln n$ can be removed.
Note that the efficiency of the algorithm depends on how fast we can compute $d_\Omega(\hmu_{i,T_i(\tau)},\hsigma^2_{i,T_i(\tau)})$: For the threshold variance problem it can be done efficiently, because $d_\Omega(x,y) = |y - v|$, i.e. it can be done in constant time.

\section{Conclusion and discussion}\label{section:discussion}
We described a framework for the risk-averse regret minimization without restriction to any particular risk measure. For a specific class of risk measures, which are functions of the mean and the variance, we proposed two algorithms that achieve logarithmic regret: one for the case of continuous functions and the one for the case of discontinuous functions. In the former case we proved logarithmic regret bound for any continuous function, while in the latter the problem need to satisfy a mild and reasonable assumption that arms should not hit the discontinuity points of the risk measure. Under this condition, the algorithms presented achieves the logarithmic regret.

We believe that assumption \ref{assumption1} might not be a necessary condition for learning. For example, even for the case when the risk measure is the Dirichlet function of the mean (which is continuous nowhere), it maybe be possible to design a sound algorithm, following the lines of \cite{Cover02}.

We remark that achieving optimal constants was not our goal and it is very likely that our bounds can be improved. An open problem, which we have not addressed in our work, is lower bounds on the risk-averse regret. Since the standard bandit problem is a particular case of our problem, we know that in this case the bound is tight (up to a constant), but obtaining a general lower bound remains an interesting research direction.
Another open problem is the extension of our results to other classes of functions. While a long-term goal would be to consider general functionals, the class of coherent risk measures could be a plausible next step. 
It is interesting to note that while classes of coherent risk measures and general functions of the mean and the variance intersect, there is no inclusion in either direction.
Finally, it is an interesting question to consider the best arm identification problem (e.g. \cite{Bubeck02}) in the context of our framework. This problem is usually referred to as a pure exploration problem, where the goal is to explore the arms in the most efficient way, focusing on minimizing the notion of simple regret.

\section{Proofs}\label{section:proofs}

\begin{proof}[Lemma \ref{lemma:continuous}]
We will prove the lemma by directly constructing a candidate function, satisfying the stated conditions.
First note that by Heine-Cantor theorem $f$ is uniformly continuous, since the domain $D$ is compact. Consider a sequence $\varepsilon_i = 2^{-i}$ for $i \geq 0$, then for every such $\varepsilon_i$ there exists $\delta_i > 0$, such that $||\bx_2-\bx_1||_1 < \delta_i \Rightarrow |f(\bx_2)-f(\bx_1)| < \varepsilon_i$ by uniform continuity. We now decrease each $\delta_i$ such that $\delta_i \leq \epsilon_i$ (if it is not the case). This does not invalidate the previous implication. 
Afterwards we construct the function $\psi$. First, $\psi(0) = 0$. Then for any $z < \delta_0$ we define
\[
k(z) = \max{\left\lbrace i : z \leq \delta_i \right\rbrace }\enspace .
\]
Then $\psi(z) = \varepsilon_{k(z)} = 2^{-k(z)}$ for $z < \delta_0$. Now we need to deal with the case when $z \geq \delta_0$. For this note that the fact $||\bx_2-\bx_1||_1 < \delta \Rightarrow |f(\bx_2)-f(\bx_1)| < \varepsilon$ for any $\bx_1,\bx_2 \in D$ implies $||\bx_2-\bx_1||_1 < 2\delta \Rightarrow |f(\bx_2)-f(\bx_1)| < 2\varepsilon$ for any $\bx_1,\bx_2 \in D$. To see this, assume the former is true and fix $\bx_1, \bx_2$ such that $||\bx_2-\bx_1||_1 < 2\delta$. Take $\bz = \frac{1}{2}\cdot (\bx_2+\bx_1)$, then for both $\bx_1$ and $\bx_2$: $||\bx_i-\bz||_1 < \delta$ and hence $|f(\bx_i)-f(\bz)| < \varepsilon$. But then
\[
|f(\bx_2)-f(\bx_1)| \leq |f(\bx_2) - f(\bz)| + |f(\bz) - f(\bx_1)| < 2\varepsilon\enspace .
\]
We use the just proven fact to define $\psi$ for $z \geq \delta_i$. Let $i$ be the smallest $i$ such that $z < 2^i \delta_0$, then $\psi(z) = 2^i \varepsilon_0$. To unify both cases we introduce
\[
a_i = \begin{cases}
 \delta_{-i} &\mbox{ if $i \leq 0$ } \\
  2^i \delta_0 &\mbox{ if $i > 0$ }\enspace .
       \end{cases}
\]
Letting $k(z) = \min{\left\lbrace i : z \leq a_i \right\rbrace }$, for $z>0$. We then have that $\psi(z) = 2^{k(z)}$. By construction, $\psi$ satisfy Condition~1 and Condition~3 of the lemma (for any $\bx_2, \bx_1 \in D:$ $ ||\bx_2-\bx_1||_1 \leq a_{k(||\bx_2-\bx_1||_1)} $, and then $|f(\bx_2)-f(\bx_1)| \leq 2^{k(||\bx_2-\bx_1||_1)} = \psi({||\bx_2-\bx_1||_1})$). 
Also, $\psi$ is well-defined, since for all $z>0$ (1)~there exists some $i$ such that $z\leq 2^i\delta_0$; and (2)~we have that $\forall i:\delta_i\leq \epsilon_i=2^{-i}$ and thus $k(z)\geq -i$ for $2^{-i}\leq z$.
To deal with Condition~2, we can take any strictly increasing function $\varphi$ that dominates $\psi$ at every point. 
For example, we can linearly interpolate between discontinuity points, i.e. define $\varphi$ as
\[
\varphi(z) = \frac{1}{a_{k(z)} - a_{k(z)-1}}\left( 2^{k(z)-1}(z - a_{k(z)-1}) + 2^{k(z)}(a_{k(z)}-z) \right) 
\]
for $z > 0$ and $\varphi(0) = 0$. It is strictly increasing (because $\psi$ is increasing, which we get from the definition of $k(z)$) and Condition~3 follows from $\psi(z) \leq \varphi(z)$ for $z \geq 0$.
\end{proof}

\begin{proof}[Theorem \ref{theorem:continuous}] The proof is similar to Theorem 1 from \cite{Sani01} with minor modifications.
We start with the following standard regret decomposition (recall that $\Delta_i = f_i - f_{i^\star}$).
\begin{equation}\label{th1:regret}
\regret_n = \sum_{t=1}^{n}f_{a_t} - \sum_{t=1}^{n}f_{i^\star} = \sum_{i: \Delta_i > 0}\Delta_i T_i(n)
\end{equation}
Hence, our task is reduced to bounding $T_i(n)$ for each arm. 
First, let $\mu_{i}^{(2)}$ be the second moment of the distribution of the arm $i$, i.e. $\mu_{i}^{(2)} = \EE{Y^2}$, where $Y \sim \nu_i$. Then
\[
\hmu_{i,t}^{(2)} = \frac{1}{t}\sum_{s=1}^{t}X_{i,s}^2
\]
is the estimator of $\mu_{i}^{(2)}$. Now we define a high probability event
\begin{equation}\label{th1:eventA}
A = \left\lbrace \forall t = 1, \dots, n; \forall i = 1, \dots, K : |\hmu_{i,t} - \mu_i| \leq \sqrt{\frac{\ln\frac{1}{\delta}}{2t}} \textrm{ and } |\hmu_{i,t}^{(2)} - \mu_{i}^{(2)}| \leq \sqrt{\frac{\ln\frac{1}{\delta}}{2t}} \right\rbrace\enspace .
\end{equation}

Using Lemma \ref{lemma:chernoff} and union bound, one can get that $\PP{A^c} \leq 4Kn\delta$. From Lemma 2 in \cite{Antos01}, we get that (\ref{bound:deviation}) holds on $A$ and, consequently, (\ref{bound:risk}) also holds on $A$ (for every $t = 1, \dots, n$ and $i = 1, \dots, K$).

Now let us consider the moment when arm $i$ is chosen at some time step $t$. It means that its lower confidence index was lower than that of the best arm (by (\ref{decision_rule})):
\begin{align*}
& f(\hmu_{i,T_i(t-1)}, \hsigma^2_{i,T_i(t-1)}) - \varphi\left(6\sqrt{\frac{\ln\frac{1}{\delta}}{2\cdot T_i(t-1)}}\right) \leq \\ & f(\hmu_{i^\star,T_i^\star(t-1)}, \hsigma^2_{i^\star,T_i^\star(t-1)}) - \varphi\left(6\sqrt{\frac{\ln\frac{1}{\delta}}{2\cdot T_i^\star(t-1)}}\right)\enspace .
\end{align*}
We also know that on the event $A$ (by (\ref{bound:risk})):
\[
f_i - \varphi\left(6\sqrt{\frac{\ln\frac{1}{\delta}}{2\cdot T_i(t-1)}}\right) \leq f(\hmu_{i,T_i(t-1)}, \hsigma^2_{i,T_i(t-1)}) 
\]
and
\[
f(\hmu_{i^\star,T_i^\star(t-1)}, \hsigma^2_{i^\star,T_i^\star(t-1)}) - \varphi\left(6\sqrt{\frac{\ln\frac{1}{\delta}}{2\cdot T_i^\star(t-1)}}\right) \leq f_{i^\star}\enspace .
\]
Combining the last three inequalities,
\[
f_i - 2\varphi\left(6\sqrt{\frac{\ln\frac{1}{\delta}}{2\cdot T_i(t-1)}}\right) \leq f_{i^\star}\enspace .
\]
Since $\varphi$ is strictly increasing function it has a well-defined inverse $\invphi$ and we can bound $T_i(t-1)$ as follows:
\[
T_i(t-1) \leq \frac{18\cdot \ln\frac{1}{\delta}}{(\invphi(\Delta_i/2))^2}\enspace .
\]
If $t$ is the last time when arm $i$ is pulled, then $T_i(n) = T_i(t-1)+1$ and hence
\begin{equation}\label{th1:final_bound}
T_i(n) \leq \frac{18\cdot \ln\frac{1}{\delta}}{(\invphi(\Delta_i/2))^2} + 1\enspace .
\end{equation}
Inserting this into (\ref{th1:regret}) gives us the stated regret bound.
\end{proof}


\begin{proof}[Theorem \ref{theorem:discontinuous}]
Again, as in Theorem \ref{theorem:continuous}, we are going to use regret decomposition (\ref{th1:regret}). Hence, we will focus on bounding $T_i(n)$ for each arm~$i$. We define the event $A$ as in (\ref{th1:eventA}) and everything we are deriving next is conditioned on $A$. We introduce the following stopping times $\lambda_i$ as
\[
\lambda_i = \inf\left\lbrace t: 6\sqrt{\frac{\ln\frac{1}{\delta}}{2t}} \leq \frac{1}{2}\cdot d_\Omega(\hmu_{i,t},\hsigma^2_{i,t}) \right\rbrace\enspace .
\]
Then we have
\[
T_i(n) = \lambda_i + \widetilde{T}_i(n)\enspace ,
\]
where $\widetilde{T}_i(n)$ is the number of times the arm $i$ was pulled during the second phase of the algorithm. Conditioned on $A$ it can be bounded as in Theorem \ref{theorem:continuous} by (\ref{th1:final_bound}) with corresponding $\varphi_i$. Next we focus on $\lambda_i$. If we define 
\[
\widetilde{\lambda}_i = \inf\left\lbrace t: 6\sqrt{\frac{\ln\frac{1}{\delta}}{2t}} \leq \frac{e_i}{3} \right\rbrace = \inf\left\lbrace t: 6\sqrt{\frac{\ln\frac{1}{\delta}}{2t}} \leq \frac{d_\Omega(\mu_i,\sigma^2_i)}{3} \right\rbrace\enspace ,
\]
then, at time $\widetilde{\lambda}_i$ Condition (\ref{stopping}) is necessarily fulfilled:
\begin{align*}
6\sqrt{\frac{\ln\frac{1}{\delta}}{2t}} &\leq \frac{d_\Omega(\mu_i,\sigma^2_i)}{3} \\
&\leq \frac{d_i(\hmu_{i,t},\hsigma^2_{i,t})}{3} + \frac{d_\Omega(\hmu_{i,t},\hsigma^2_{i,t})}{3} \\
&\leq \frac{1}{3} \cdot 6\sqrt{\frac{\ln\frac{1}{\delta}}{2t}} + \frac{d_\Omega(\hmu_{i,t},\hsigma^2_{i,t})}{3}\enspace .
\end{align*}
Hence $\lambda_i \leq \widetilde{\lambda}_i = 162\cdot e_i^{-2}\ln\frac{1}{\delta}$. Combining this together with (\ref{th1:final_bound}) and (\ref{th1:regret}) gives the stated result.
\end{proof}

\bibliographystyle{splncs03}
\bibliography{biblio}{}

\end{document}